\documentclass[conference]{IEEEtran}
\usepackage{times}
\usepackage{cite}

\usepackage{ upgreek }
\usepackage{verbatim} 

\usepackage{amssymb}
\usepackage{graphicx}
\usepackage{url}
\usepackage{verbatim} 
\usepackage{color}
\usepackage{wrapfig}

\usepackage{array}
\usepackage{relsize}
\usepackage{multirow}
\usepackage{amsmath}
\usepackage[numbers]{natbib}
\usepackage{multicol}
\usepackage[bookmarks=true]{hyperref}

\DeclareMathOperator*{\argmax}{argmax}
  
\newlength\savedwidth
\newcommand\whline[1]{\noalign{\global\savedwidth\arrayrulewidth
                               \global\arrayrulewidth #1} %
                      \hline
                      \noalign{\global\arrayrulewidth\savedwidth}}

\newcommand{\todo}[1]{\textcolor{blue}{\textbf{#1}}}

\newcommand{\x}{{\mathbf x}}     
\newcommand{\xs}[1]{{x_{#1}}}    
\newcommand{\y}{{\mathbf y}}     
\newcommand{\ys}[1]{{y_{#1}}}    
\newcommand{\ysc}[2]{{y_{#1}^{#2}}}    
\newcommand{\zsc}[2]{{z_{#1}^{#2}}}    
\newcommand{\fn}[1]{{\phi_n(#1)}}      
\newcommand{\fe}[3]{{\phi_{#1}(#2,#3)}}
\newcommand{\w}{{\mathbf w}}           
\newcommand{\wn}[1]{{w_n^{#1}}}        
\newcommand{\we}[3]{{w_{#1}^{#2#3}}}   
\newcommand{\df}[3]{{f_{#3}(#1,#2)}}   

\newenvironment{packed_item}{
\begin{itemize}
  \setlength{\itemsep}{1pt}
  \setlength{\parskip}{0pt}
  \setlength{\parsep}{0pt}
}{\end{itemize}}

\newlength{\sectionReduceTop}
\newlength{\sectionReduceBot}
\newlength{\subsectionReduceTop}
\newlength{\subsectionReduceBot}
\newlength{\abstractReduceTop}
\newlength{\abstractReduceBot}
\newlength{\captionReduceTop}
\newlength{\captionReduceBot}
\newlength{\subsubsectionReduceTop}
\newlength{\subsubsectionReduceBot}

\newlength{\horSkip}
\newlength{\verSkip}

\newlength{\figureHeight}
\long\def\symbolfootnote[#1]#2{\begingroup%
\def\thefootnote{\fnsymbol{footnote}}\footnotetext[#1]{#2}\endgroup}

\begin{document}

\title{Labeling 3D scenes for Personal Assistant Robots}

\author{Hema Swetha Koppula$^*$, Abhishek Anand$^*$, Thorsten Joachims, Ashutosh Saxena\\
Department of Computer Science, Cornell University.\\
\texttt{\{hema,aa755,tj,asaxena\}@cs.cornell.edu}}



%

\maketitle

\begin{abstract}

Inexpensive RGB-D cameras that give an RGB image together with depth data 
have become widely available. We use this data to build 3D point clouds of
a full scene.
In this paper, we address the task of labeling objects in this 3D point cloud of a 
complete indoor scene such as an office. We propose a graphical model that captures 
 various features and contextual relations, including the local visual appearance and 
 shape cues, object co-occurrence relationships and geometric relationships.  With a
 large number of object classes and relations, the model's parsimony becomes 
 important and we address that by using multiple types of edge potentials. The model
 admits efficient approximate inference, and we train it using a maximum-margin 
 learning approach.  In our experiments over a total
of 52 3D scenes of homes and offices (composed from about 550 views, having
2495 segments labeled with 27 object classes), we get a performance of 
84.06\% in labeling 17 object classes for offices, and 73.38\%
in labeling 17 object classes for home scenes. Finally, we applied these algorithms
successfully on a mobile robot for the task of finding an object in a large cluttered room.

\end{abstract}

\symbolfootnote[0]{$^*$ indicates equal contribution.}
\section{Introduction}
\vspace{\sectionReduceBot}

Inexpensive RGB-D sensors that augment an RGB image with depth data have recently become widely available. At the same time, years of research on SLAM (Simultaneous Localization and Mapping) now
make it possible to 
merge multiple RGB-D images into a single point cloud, easily providing an approximate 3D model of a complete indoor scene (e.g., a room). In this paper, we explore how this move from part-of-scene 2D images to full-scene 3D point clouds can improve the richness of models for object labeling.



In the past, a significant amount of work has been done in semantic labeling of 2D images. However, a 
lot of valuable information about the shape and geometric layout of objects is lost when a 2D image is 
formed from the corresponding 3D world.   A classifier that has access to a full 3D model, can access 
important geometric properties in addition to the local shape and appearance of an object.  For example, 
many objects occur in characteristic relative geometric configurations (e.g., a monitor is almost always on a table), and many objects consist of visually distinct parts that occur in a certain relative configuration. More generally, a 3D model makes it easy to reason about a variety of properties, which are based on 3D distances,
volume and local convexity.

\begin{figure*}[t!]
 \centering
 \includegraphics[width=0.5\linewidth,height=0.15in]{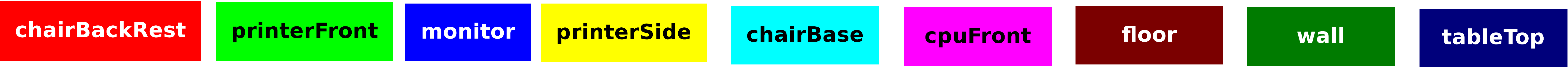}\\
\includegraphics[width=.32\linewidth,height=0.65in]{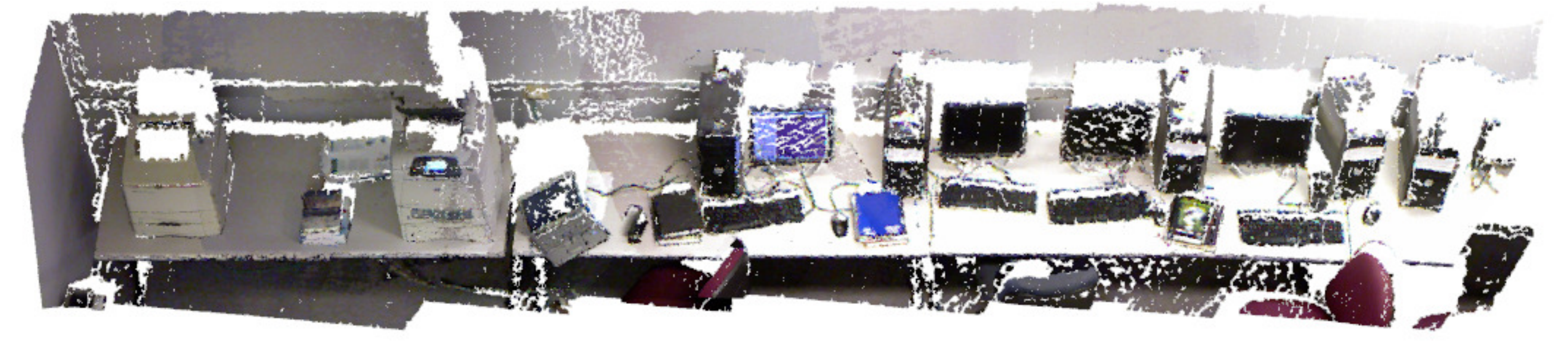} 
\includegraphics[width=.32\linewidth,height=0.65in]{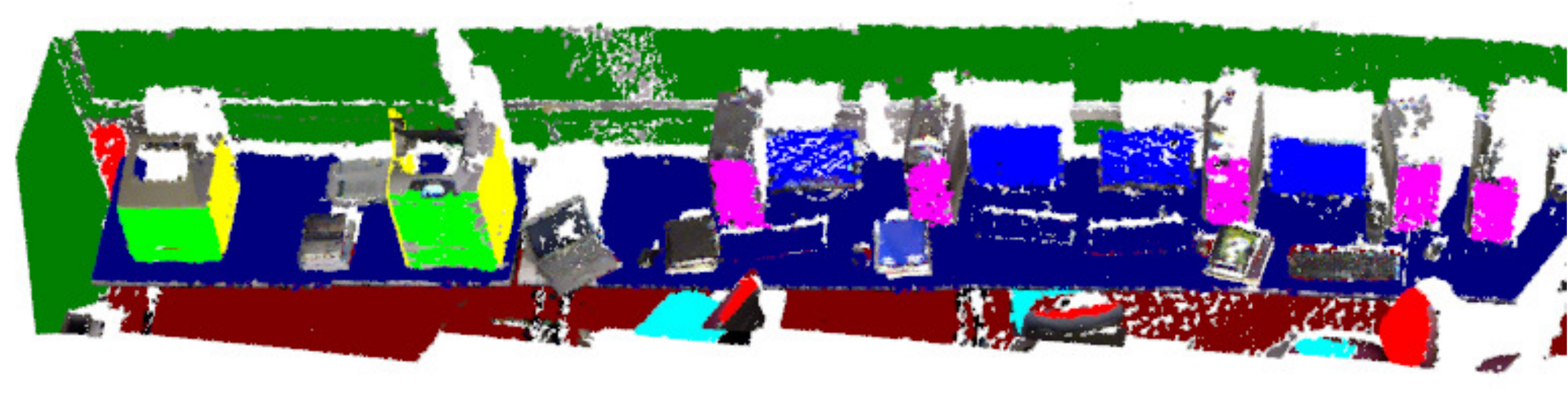} 
\includegraphics[width=.32\linewidth,height=0.65in]{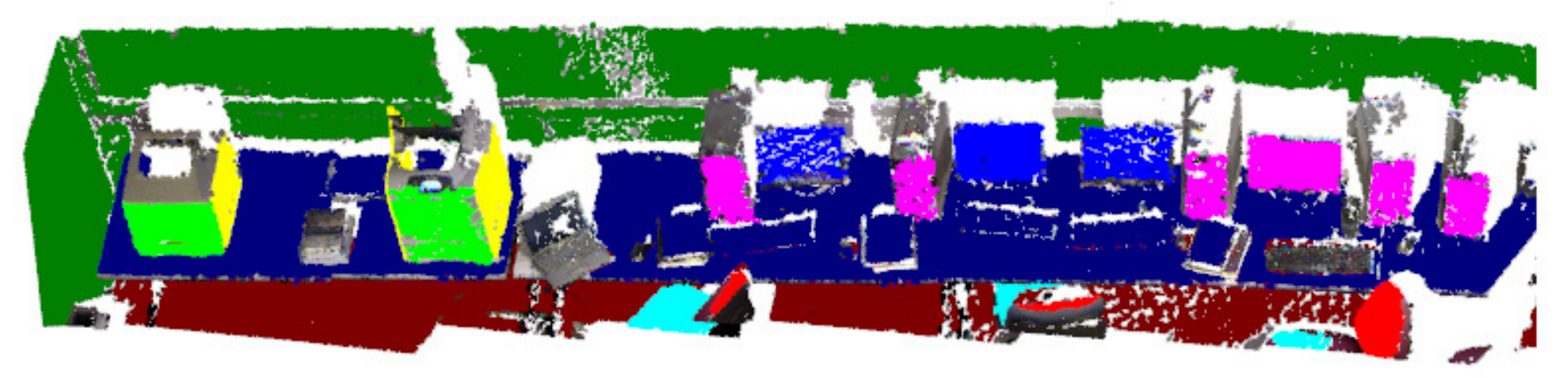} \\
 \includegraphics[width=0.5\linewidth,height=0.15in]{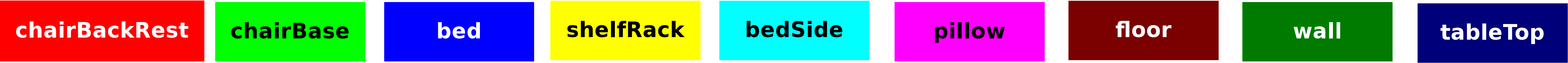}\\
\includegraphics[width=.32\linewidth,height=0.75in]{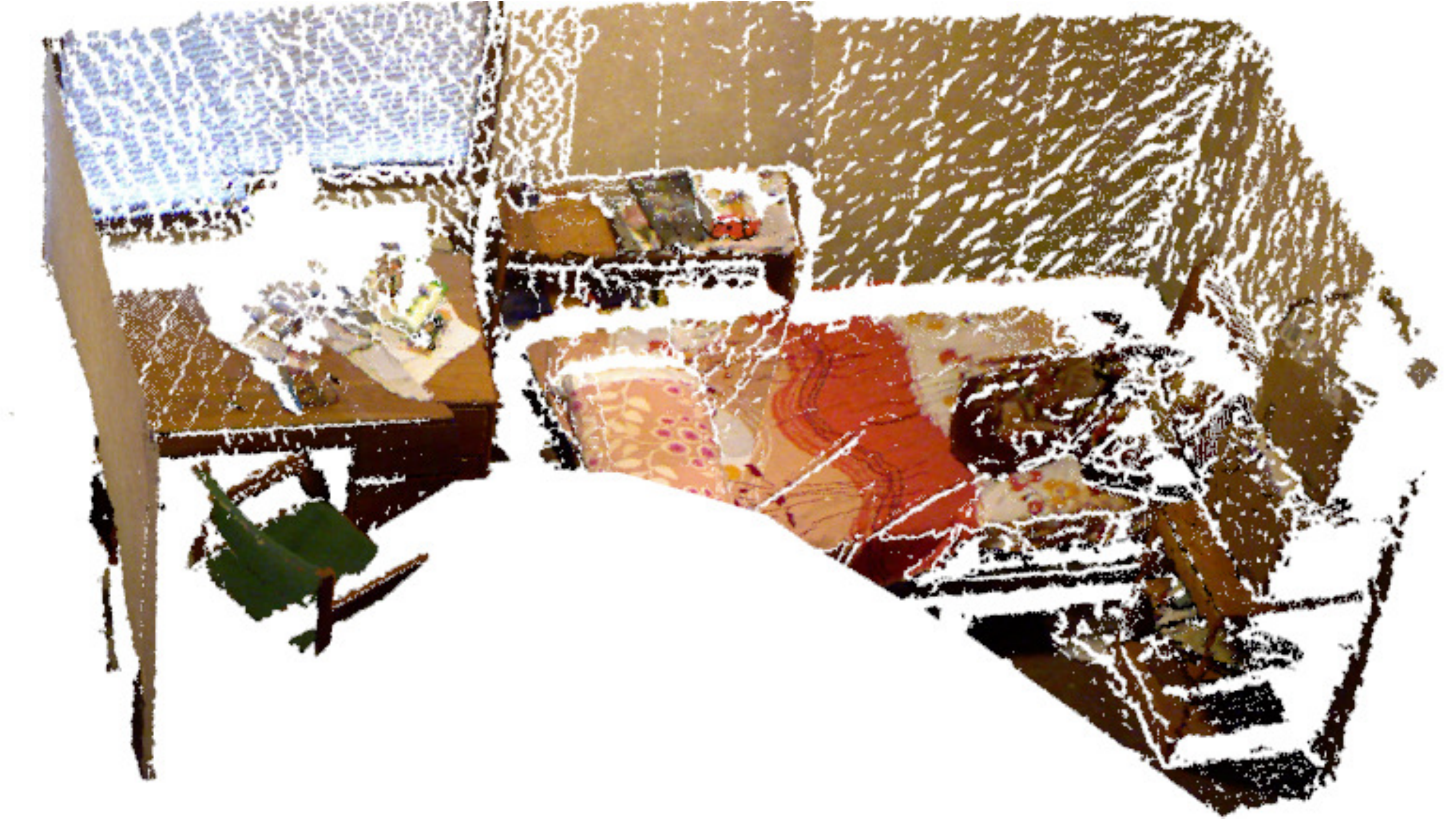} 
\includegraphics[width=.32\linewidth,height=0.75in]{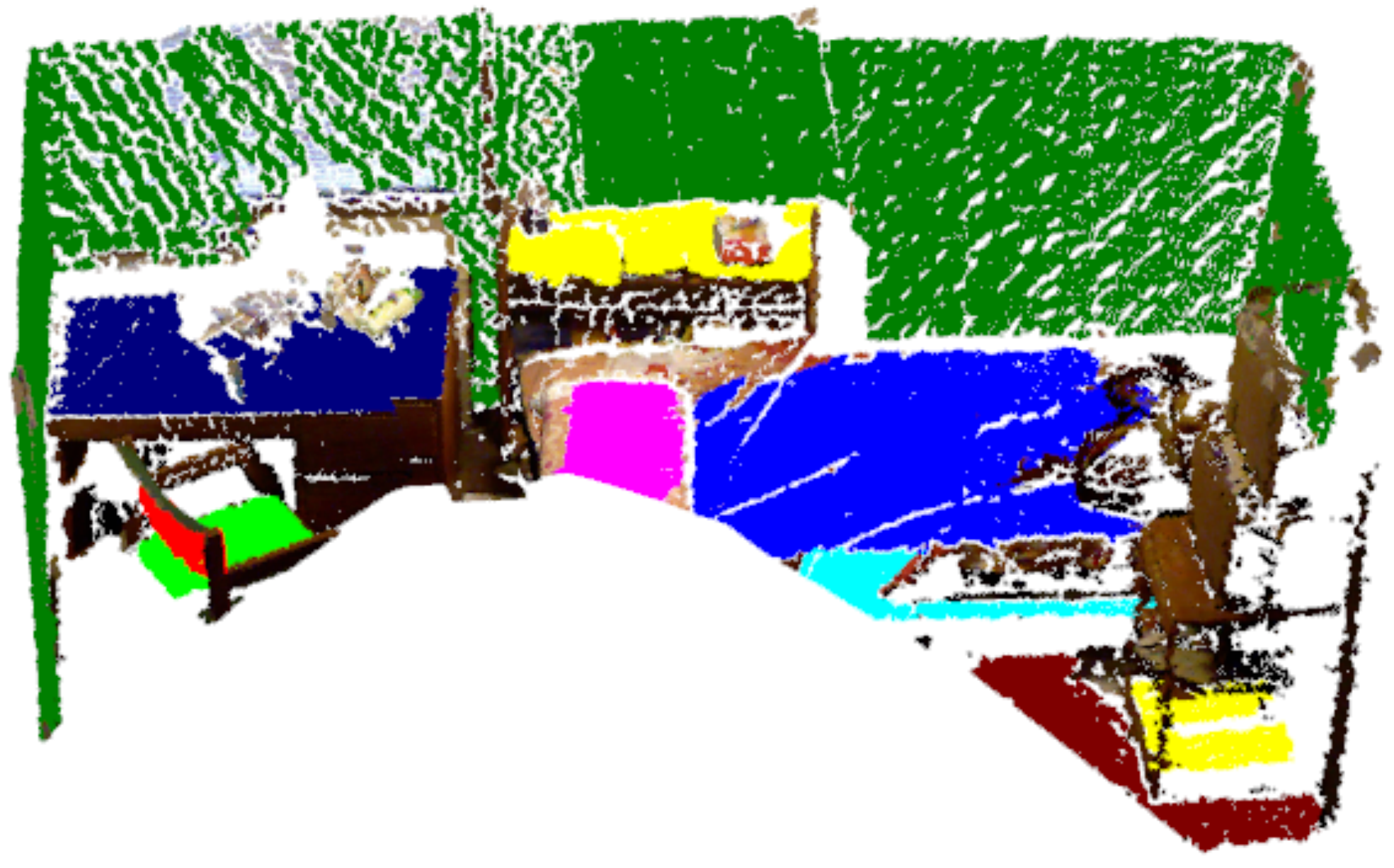} 
\includegraphics[width=.32\linewidth,height=0.75in]{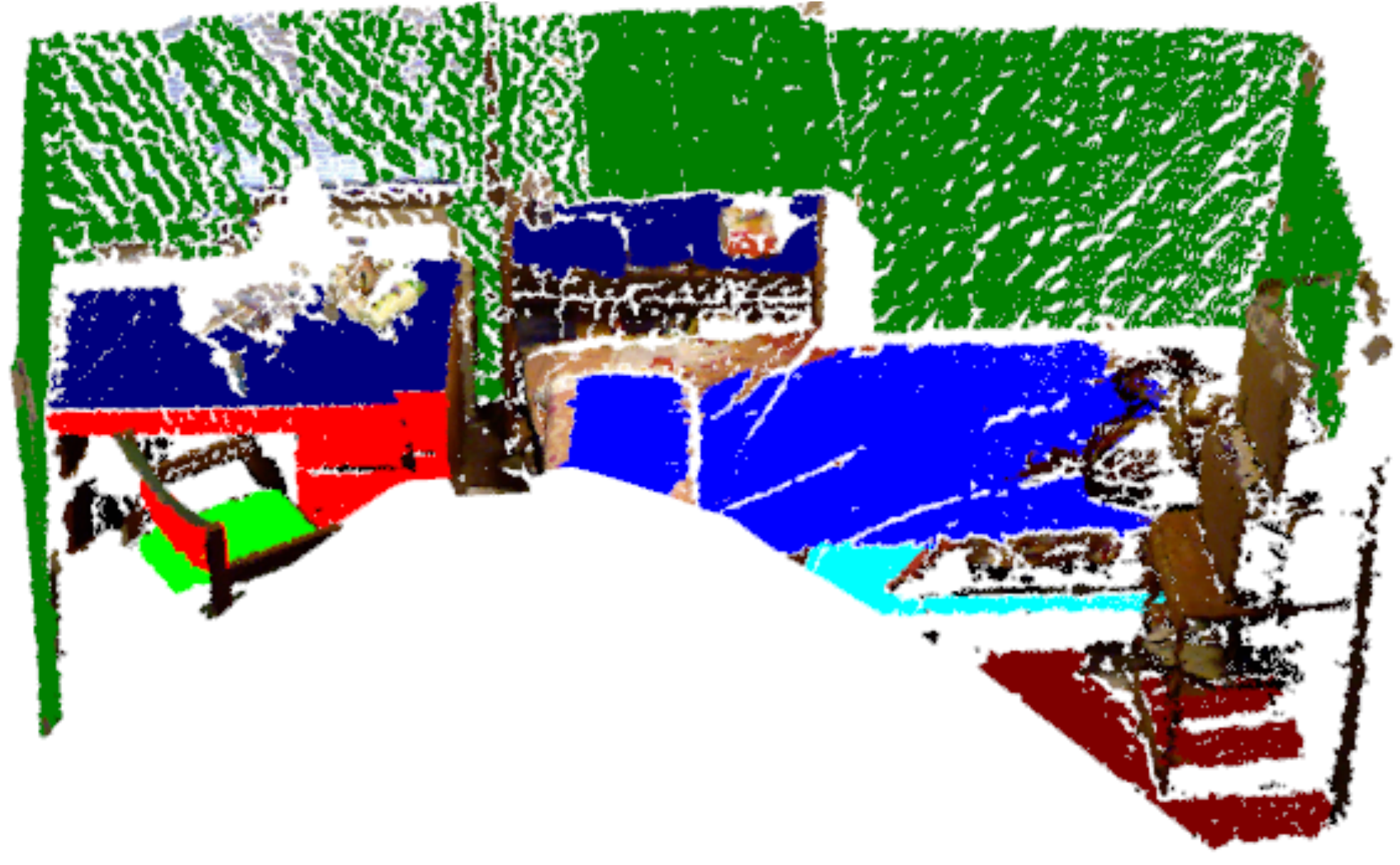} 
\vspace*{\captionReduceTop}
 \caption{{\small Office scene (top) and Home scene (bottom with the corresponding label coloring above the images. The left-most is the original point cloud, the middle is the ground truth labeling and the right most is the point cloud with predicted labels.}}
\label{fig:examplePCD}
\vskip -.1in
 \end{figure*}

In our work, we first use SLAM in order to compose multiple views from a 
Microsoft Kinect RGB-D sensor together into one 3D point cloud, providing each RGB pixel 
with an absolute 3D location in the scene. We then (over-)segment the scene and predict 
semantic labels for each segment (see Fig.~\ref{fig:examplePCD}). We predict not only coarse 
classes like in \cite{xiong:indoor,Anguelov/etal/05} (i.e., wall, ground, ceiling, building), but also 
label individual objects (e.g., printer, keyboard, mouse). Furthermore, we model rich relational 
information beyond an associative coupling of labels \cite{Anguelov/etal/05}.

In this paper, we propose and evaluate the first model and learning algorithm for scene understanding that exploits rich relational information derived from the full-scene 3D point cloud for object labeling. In particular, we propose a graphical model that naturally captures the geometric relationships of a 3D scene. Each 3D segment is associated with a node, and pairwise potentials model the relationships between segments (e.g., 
co-planarity, convexity, visual similarity, object occurrences and proximity). The model admits efficient approximate inference \cite{Kolmogorov/Rother/07}, and we show that it can be trained using a maximum-margin approach \cite{Taskar/AMN,Tsochantaridis/04,Finley/Joachims/08a} that globally minimizes an upper bound on the training loss. We model both associative and non-associative  coupling of labels.
With a large number of object classes, the model's parsimony becomes important. 
Some features are better indicators of label similarity, while other features
are better indicators of non-associative relations such as geometric arrangement (e.g., ``on top
of," ``in front of"). We therefore model them using appropriate clique potentials rather than using general clique potentials.
 Our model is highly flexible and we 
 have made
 our software available for download to other researchers in this emerging 
area of 3D scene understanding. 


To empirically evaluate our model and algorithms, we perform several experiments over a total of 
52 scenes of two types: offices and homes.  These scenes were built from about 550 views 
from the Kinect sensor, and they will also be made available for public use. We consider labeling 
each segment (from a total of about 50 segments per scene) into  27 classes (17 for offices and 
17 for homes, with 7 classes in common). Our experiments show that our method, which captures 
several local cues and contextual properties, achieves an overall performance of 84.06\% on office 
scenes and 73.38\% on home scenes. We also consider the problem of labeling 3D segments 
with multiple attributes meaningful to robotics context (such as small objects that can be manipulated, 
furniture, etc.). Finally, we successfully applied these algorithms on a mobile robot for the task of 
finding an object  in a large cluttered room.

\vspace*{\sectionReduceTop}
 \section{Related Work}
 \vspace*{\sectionReduceBot}
 
There is a huge body of work in the area of scene understanding and object recognition from 2D images. 
Previous works focus on several different aspects: designing good local features such as 
HOG (histogram-of-gradients)  \cite{dalal2005histograms} and bag of words \cite{csurka2004visual},  designing good global (context) features such as GIST features \cite{GIST}, 
and combining multiple tasks \cite{li2010feedback}.
However, these approaches
do not consider the relative arrangement of the parts of the object or of multiple objects with 
respect to each other.  A number of works
propose models that explicitly capture the relations between different parts of the object
\cite{felzenszwalb2008discriminatively}, and between different objects in 2D images 
 \cite{Torralba:exploting_context,HeitzECCV_usingstufftofindthings}.
 However, a lot of valuable information about the shape and geometric layout of objects is lost when a 2D image is formed from the corresponding 3D world.  In some recent works, 3D layout or depths have been used for improving object detection (e.g.,  
 \cite{saxena-depth,Hoiem:puttingobjects,saxena-make3d-pami,hebert-room-layout,Leibe07:dynamic}). 
 Here a rough 3D scene geometry (e.g., the main surfaces in a scene)
  is inferred from a single 2D image or a stereo video, respectively. 
However, the estimated geometry is not accurate enough to give significant 
improvements.
 With 3D data, we can more precisely determine the shape, size and geometric orientation of the objects, 
 and several other properties and therefore capture much stronger context.
 The recent availability of synchronized videos of both color and depth obtained from RGB-D (Kinect-style) 
depth cameras, shifted the focus to making use of both visual as well as 
shape features for  object detection \cite{gould:fusion,quigley:high-accuracy,Rusu:ObjectMaps,lai:icra11a, lai:icra11b} and 3D segmentation   
(e.g.,~\cite{Collet:icra2011}).
These methods demonstrate that augmenting 
visual features with 
3D information can enhance object detection in cluttered, real-world environments. 
 However, these works do not make use of the contextual 
relationships between various objects which have been shown to be useful for tasks such as object 
detection and scene understanding in 2D images.
Our goal is
to perform  semantic labeling of indoor 3D scenes by modeling and learning several contextual
relationships. 


  There is also some recent work in labeling outdoor scenes obtained from LIDAR data into
  a few geometric classes (e.g., ground, building, trees, vegetation, etc.).  \cite{Golovinskiy:shape-basedrecognition,Shapovalov2010} capture context by designing node features and \cite{xiong:3DSceneAnalysis} do so by stacking layers of classifiers; however these methods do not
model the correlation between the labels.  
Some of these works model some contextual relationships in the learning model itself. For example, \cite{Munoz2009:mrf,Anguelov/etal/05} use 
  associative Markov networks in order to favor similar labels for nodes in the cliques.
However, many relative features between objects are not associative in nature. 
For example, the relationship ``on top of" does not hold in between two ground segments, i.e., a ground segment cannot be ``on top of" another ground segment. Therefore, using an associative Markov network is very restrictive for our problem.  
All of these works \cite{Shapovalov2010,Munoz2009:mrf,Anguelov/etal/05,xiong:3DSceneAnalysis} were designed for outdoor scenes with LIDAR data (without RGB values) and therefore would not apply directly to RGB-D data in indoor environments.
Furthermore, these methods only consider very few geometric classes (between three to five classes) in outdoor environments, whereas we consider a large number of object classes for labeling the indoor RGB-D data.

The most related work to ours is \cite{xiong:indoor}, where they label the planar patches in a point-cloud of 
an indoor scene with four geometric labels (walls, floors, ceilings, clutter).  They use a CRF to 
 model geometrical relationships such as orthogonal, parallel, adjacent, and coplanar.
 The learning method for estimating the parameters was based on maximizing the pseudo-likelihood
 resulting in a sub-optimal learning algorithm.
In comparison, our basic representation is 3D segments (as compared to planar patches) and 
we consider a much larger number of classes (beyond just the geometric classes).
We capture a much richer set of relationships between pairs of objects, and
use a principled max-margin learning method to learn the parameters of our model.

\vspace*{\sectionReduceTop}
\section{Approach}\label{S.properties}
\vspace*{\sectionReduceBot}

We now outline our approach, including the model, its inference methods, and the learning algorithm. 
Our input is multiple Kinect RGB-D images of an indoor scene stitched into a single 
3D point cloud using RGBDSLAM \cite{rgbdslam}.
Each such point cloud is 
then over-segmented based on smoothness (i.e., difference in the local surface normals) and 
continuity of surfaces  (i.e.,  distance between the points). These segments are the 
atomic units in our model. Our goal is to label each of them.


Before getting into the technical details of the model, the following outlines the properties we aim 
to capture:



\noindent
\textbf{Visual appearance.} The reasonable success of object detection in 2D images 
shows that visual appearance is a good indicator for labeling scenes.
We therefore model the local color, texture, gradients of intensities, etc.\ 
for predicting the labels.
In addition, we also model the property that if nearby segments are similar in 
visual appearance, they are more likely to belong to the same object.

\noindent
\textbf{Local shape and geometry.}  Objects have characteristic 
shapes---for example, a table is horizontal, a monitor is vertical, a keyboard is uneven, 
and a sofa is usually smoothly curved. 
Furthermore, parts of an object often form a convex shape. We compute
3D shape features to capture this.

\noindent 
\textbf{Geometrical context.}
Many sets of objects occur in characteristic relative geometric configurations. 
For example, a monitor is always \emph{on-top-of} a table, chairs are usually found \emph{near} tables, a keyboard is \emph{in-front-of} a monitor.
This means that our model needs to capture {\em non-associative} relationships 
(i.e., that neighboring segments differ in their labels in specific patterns). 

\smallskip
Note that the examples given above are just illustrative. For any particular practical application, 
there will likely be other properties that could also be included. As demonstrated in the 
following section, our model is flexible enough to include a wide range of features.

\begin {comment}
\item \textbf{Visual appearance.}  We obtain the RGB values for the points in the segment. The
color and texture of a segment is often a good indicator of what object it is.  
We used histogram-of-gradients (HOG) features that have been successfully used in many object 
recognition tasks for images \cite{dalal2005histograms,felzenszwalb2008discriminatively}. 
We also compute average hue, saturation and intensity values of a segment and their histograms. 
\item \textbf{Local shape.}  Different objects have different shapes---for example, a table is flat while a keyboard
is uneven. We incorporate several shape properties based on surface normals, and other statistics
of the point-cloud in the segment. We describe these in Section~\ref{sec:features}.
\item \textbf{Co-occurence of the objects.}  Objects 
such as monitor and keyboard, or pillow and bed, usually co-occur. We need to capture these 
co-occurrence statistics in our model.
\item \textbf{Geometric properties.}  Certain objects follow strong relative location preferences,
such as ``infront-of" or ``on-top-of" 
For example, a monitor is almost always \emph{on-top-of} 
a table and a keyboard is \emph{infront-of} a monitor.  
We model several geometric relations between a pair of
 segments:  relative depth-ordering (e.g., ``infront-of", ``behind"), relative horizontal-ordering 
 (``above", ``below" and ``same height"), coplanar-ness, etc.   See Section~\ref{sec:relfeatures}
 for more details.
\end{comment}


\vspace*{\subsectionReduceTop}
\subsection{Model Formulation}
\vspace*{\subsectionReduceBot}
\label{sec:model}
We model the three-dimensional structure of a scene using a model isomorphic to a Markov Random Field with log-linear node and pairwise edge potentials. Given a segmented point cloud $\x=(\xs{1},...,\xs{N})$ consisting of segments $\xs{i}$, we aim to predict a labeling $\y=(\ys{1},...,\ys{N})$ for the segments. Each segment label $\ys{i}$ is itself a vector of $K$ binary class labels $\ys{i}=(\ysc{i}{1},...,\ysc{i}{K})$, with each $\ysc{i}{k} \in \{0,1\}$ indicating whether a segment $i$ is a member of class $k$. Note that multiple $\ysc{i}{k}$ can be $1$ for each segment (e.g., a segment can be both a ``chair'' and a ``movable object''). We use such multi-labelings in our attribute experiments where each segment can have multiple attributes, but not in segment labeling experiments where each segment can have only one label. 

For a segmented point cloud $\x$, the prediction $\hat{\y}$ is computed as the argmax of a discriminant function $\df{\x}{\y}{\w}$ that is parameterized by a vector of weights $\w$.
\begin{equation} \label{eq:argmax}
\hat{\y} = \argmax_\y \df{\x}{\y}{\w}
\end{equation}
The discriminant function captures the dependencies between segment labels as defined by an undirected graph $(\mathcal{V},\mathcal{E})$ of vertices $\mathcal{V} = \{1,...,N\}$ and edges $\mathcal{E} \subseteq 
\mathcal{V} \times \mathcal{V}$. We describe in Section~\ref{sec:features} how this graph is derived from the spatial proximity of the segments.
 Given $(\mathcal{V},\mathcal{E})$, we define the following discriminant function based on individual segment features $\fn{i}$ and edge features $\fe{t}{i}{j}$ as further described below.
\small
\begin{eqnarray} \label{eq:model}
\df{\y}{\x}{\w} = \sum_{i \in \mathcal{V}} \sum_{k=1}^{K} \ysc{i}{k} \left[\wn{k} \cdot \fn{i} \right] \nonumber \\
+  \sum_{(i,j)\in \mathcal{E}}   \sum_{T_t \in {\cal T}}  \sum_{(l,k)\in T_t} \ysc{i}{l} \ysc{j}{k}  \left[\we{t}{l}{k} \cdot \fe{t}{i}{j}\right] 
\end{eqnarray}
\normalsize
The node feature map $\fn{i}$ describes segment $i$ through a vector of features, and there is one weight vector for each of the $K$ classes. Examples of such features are the ones capturing local visual appearance, shape and geometry. The edge feature maps $\fe{t}{i}{j}$ describe the relationship between segments $i$ and $j$. Examples of edge features are the ones capturing similarity in visual appearance and geometric 
context.\footnote{Even though it is not represented in the notation, note that both the node feature map $\fn{i}$ and the edge feature maps $\fe{t}{i}{j}$ can compute their features based on the full $\x$, not just $\xs{i}$ and $\xs{j}$.}
There may be multiple types $t$ of edge feature maps $\fe{t}{i}{j}$, and each type has a graph over the $K$ classes with edges $T_t$. If $T_t$ contains an edge between classes $l$ and $k$, then this feature map and a weight vector $\we{t}{l}{k}$ is used to model the dependencies between classes $l$ and $k$. If the edge is not present in $T_t$, then $\fe{t}{i}{j}$ is not used.

We say that a type $t$ of edge features is modeled by an associative edge potential if ${T_t}=\{(k,k)| \forall k=1..K\}$. And it is modeled by an non-associative edge potential if $T_t=\{(l,k)| \forall l,k=1..K\}$. Finally, it is modeled by an object-associative edge potential if $T_t=\{(l,k) | \exists object , ~ l,k\in {\rm parts}({\rm object})\}$\\
 
\noindent
\textbf{Parsimonious model.} 
In our experiments we distinguished between two types of edge feature maps---``object-associative'' features $\fe{oa}{i}{j}$ used between classes that are parts of the same object (e.g., ``chair base'', ``chair back'' and ``chair back rest''), and ``non-associative'' features $\fe{na}{i}{j}$ that are used between any pair of classes. Examples of features in the object-associative feature map $\fe{oa}{i}{j}$ include similarity in appearance, co-planarity, and convexity---i.e., features that indicate whether two adjacent segments belong to the same class or object.
 A key reason for distinguishing between object-associative and non-associate features is parsimony of the model.  In this parsimonious model (referred to as svm\_mrf\_parsimon), we model object associative features using object-associative edge potentials and non-associative features as non-associative edge potentials. As not all edge features are ``non-associative'', we avoid learning weight vectors for relationships which do not exist. Note that $|T_{na}| >> |T_{oa}|$ since, in practice, the number of parts of an objects is much less than K. Due to this, the model we learn with both type of edge features will have much lesser number of parameters compared to a model learnt with all edge features as ``non-associative features''. 

\vspace*{\subsectionReduceTop}
\subsection{\label{sec:features}Features}
\vspace*{\subsectionReduceBot}

Table \ref{tab:Features}  summarizes the features used in our experiments. ${\lambda_i}_0,{\lambda_i}_1$ and ${\lambda_i}_2$ are the 3 eigen-values of the scatter matrix computed from the points of segment $i$ in increasing order. $c_i$ is the centroid of  segment $i$. $r_i$ is the ray vector to the $c_i$ from the camera in which it was captured. $rh_i$ is the projection of $r_i$ on horizontal plane. 
$\hat{n}_i$ is the unit normal of segment i which points towards the camera ($r_i . \hat{n}_i <0$).

The node features $\phi_n(i)$
consist of visual appearance features based on histogram of HSV values and the histogram of gradients
(HOG), as well as local shape and geometry features that capture properties such as how planar
a segment is, its absolute location above ground, and its shape. Some features capture
spatial location of an object in the scene (e.g.,  N9).

We connect two segments (nodes) $i$ and $j$ by an edge if there exists a point in segment $i$ and 
a point in segment $j$ which are less than \emph{context\_range} distance apart. This captures the
closest distance between two segments (as compared to centroid distance between the segments)---we 
study the effect of context range more in Section~\ref{sec:experiments}.
The edge features $\phi_t(i,j)$ (Table \ref{tab:Features}-right) consist of associative features (E1)
based on visual appearance and local shape, as well as non-associative features (E3-E8) that capture
the tendencies of two objects to occur in certain configurations.
Note that our features are insensitive to horizontal translation and 
rotation of the camera. However, our features place a lot of emphasis on the vertical direction 
because gravity influences the shape and relative positions of objects to a large extent.

\begin{table}[t!]
\vskip -.1in
\caption{}
\vspace*{\captionReduceBot}
\label{tab:Features}
{\scriptsize
\begin{minipage}[b]{7cm}\centering

\begin{tabular}{|p{7.0cm}|c|}
\multicolumn {2}{c}{Node features for segment $i$.} \\ 
\hline 
 Description  & Count\\ \hline
{\bf Visual Appearance} & {\bf 48} \\
N1. Histogram of HSV color values & 14 \\
N2. Average HSV color values & 3 \\
N3. Average of HOG features of the blocks in image spanned by the points of a segment & 31 \\
\hline
{\bf Local Shape and Geometry} & {\bf 8} \\
N4. linearness (${\lambda_{i0}}$ - ${\lambda_{i1}}$), planarness (${\lambda_{i1}}$ - ${\lambda_{i2}}$),Scatter: ${\lambda_i}_0$ & 3 \\ 
N6. Vertical component of the normal: ${\hat{n_i}}_z$ & 1 \\ 
N7. Vertical position of centroid: ${c_i}_z$ & 1 \\ 
N8. Vert.~and Hor.~extent of bounding box  & 2 \\ 
N9. Dist.~from the scene boundary 
& 1 \\
\hline
\end{tabular}
\end{minipage}
\hspace{0.5cm}
\begin{minipage}[b]{7cm}
\centering
\begin{tabular}{|p{7.0cm}|c|}
\multicolumn {2}{c}{Features for edge (segment $i$, segment $j$).}\\ \hline
{\bf Visual Appearance (associative)} & {\bf 3} \\
E1. Difference of avg HSV color values & 3\\
\hline
{\bf Local Shape and Geometry (associative)} & {\bf 2} \\
E2. Coplanarity and convexity 
& 2\\ 
\hline
{\bf Geometric context (non-associative)} & {\bf 6} \\

E3. Horizontal distance b/w centroids.
& 1\\ 

E4. Vertical Displacement b/w centroids:
 $(c_{iz} - c_{jz})$ 
& 1\\ 

E5. Angle between normals (Dot product):
$ \hat{n}_i \cdot \hat{n}_j $
& 1\\ 

E6. Diff.~in angle with vert.:
$ \cos^{-1}({n_{iz}})$ - $\cos^{-1}({n_{jz}}) $
& 1\\

E7. Dist.~between closest points 
& 1\\
 
E8. rel.~position from camera (in front of/behind).
& 1\\ 

\hline

\end{tabular}

\end{minipage}
}
\vskip -.2in
\end{table}

\begin {comment} 
\vspace*{\subsectionReduceTop}
\subsection{Relative Features}
\vspace*{\subsectionReduceBot}

\todo{This section describes the relative features in more detail.}

We model the following geometric relations between a pair of
 segments:  relative depth-ordering, relative horizontal-ordering, coplanar-ness, etc.  
For any given pair of segments, the relative depth-ordering  captures the ``infront-of", ``behind" and 
``none" relations and the relative  horizontal-ordering captures the ``above", ``below" and ``same height" relations.
\end{comment}

\vspace*{\subsectionReduceTop}
\subsection{Learning and Inference}
\vspace*{\subsectionReduceBot}

Solving the argmax in Eq.~\ref{eq:argmax} for the discriminant function in Eq.~\ref{eq:model} is NP hard. It can be formulated as the following mixed-integer program, which can be solved by a general-purpose MIP solver\footnote{http://www.tfinley.net/software/pyglpk/readme.html} in about 20 minutes on a typical scene. 
\small
\begin{eqnarray}
\hat{\y}\!\!\!=\!\!\!\argmax_{\y}\max_{\mathbf z} \sum_{i \in \mathcal{V}} \sum_{k=1}^{K} \ysc{i}{k} \left[\wn{k} \cdot \fn{i} \right] \nonumber\\
+\!\!\!\sum_{(i,j)\in \mathcal{E}}  \sum_{T_t \in {\cal T}} \sum_{(l,k)\in T_t} \zsc{ij}{lk} \left[\we{t}{l}{k} \cdot \fe{t}{i}{j}\right] 
 \label{eq:relaxobj}\\
 \forall i,j,l,k: \:\: \zsc{ij}{lk}\le \ysc{i}{l}, 
\zsc{ij}{lk}\le \ysc{j}{k},\:\:\:\: \ysc{i}{l} + \ysc{j}{k} \le \zsc{ij}{lk}+1 \nonumber \\
\zsc{ij}{lk},\ysc{i}{l} \in \{ 0,1 \}, \:\:\:\:
\forall i: \sum_{j=1}^{K} \ysc{i}{j} = 1 \label{eq:sum1const}
\end{eqnarray}
\normalsize
However, if we remove the last constraint (\ref{eq:sum1const}), and relax the variables $\zsc{ij}{lk}$ and $\ysc{i}{l}$ to the interval $[0,1]$, we get a linear relaxation that can be shown to always have half-integral solutions (i.e. $\ysc{i}{l}$ only take values $\{0,0.5,1\}$ at the solution) \cite{hammer1984roof}. Furthermore, this relaxation can also be solved as a quadratic pseudo-Boolean optimization problem using a graph-cut method \cite{Kolmogorov/Rother/07}, which is orders of magnitude faster than using a general purpose LP solver (i.e., 2 sec for labeling a typical full scene in our experiments, and 0.2 sec for a single view). For training, we use the $SVM^{struct}$\footnote{http://svmlight.joachims.org/svm\_struct.html} software which uses the cutting plane method to jointly learn values of $w_n$ and $w_t$'s so as to minimize a regularized upper bound on the training error.

\vspace*{\sectionReduceTop}
   \section{Experiments\label{sec:experiments}}
   
  \vspace*{\subsectionReduceTop}  
   \subsection{Data}
  \vspace*{\subsectionReduceBot}
 
  We consider labeling object segments in full 3D scene
(as compared to 2.5D data from a single view).
  For this purpose, we collected data of 24 office and 28 home scenes. Each
scene was reconstructed from about 8-9 RGB-D views from a Microsoft
Kinect sensor and we have a total of about 550 views. Each scene contains about a million colored points.
We first over-segment the 3D scene (as described earlier)
to obtain the atomic units of our representation.
For training, we manually labeled the segments, and we selected 
the labels which were present in
a minimum of 5 scenes in the dataset.  
Specifically, the office labels are: \{\textit{wall, floor, tableTop, tableDrawer,
tableLeg, chairBackRest,  chairBase, chairBack, monitor,
printerFront, printerSide keyboard,  cpuTop, cpuFront, cpuSide, book, paper}\},
and the home labels are: \{\textit{wall, floor, tableTop, tableDrawer, tableLeg, chairBackRest, 
 chairBase,  sofaBase, sofaArm, sofaBackRest, bed, bedSide, quilt, pillow, shelfRack, laptop, book}\}. This gave us a total of 1108 labeled segments in the office scenes and 1387 segments in the home scenes. 
 Often one object may be divided 
into multiple segments because of over-segmentation.
We have made this data available at: {\texttt{http://pr.cs.cornell.edu/sceneunderstanding}.




   \vspace*{\subsectionReduceTop}
   \subsection{Results}
   \vspace*{\subsectionReduceBot}
   
Table \ref{tbl:overall_result} shows the results, performed using 4-fold
cross-validation and averaging performance across the folds for the models trained 
separately on home and office datasets.  
We use both the macro and micro averaging to aggregate precision and recall
over various classes.  Since our algorithm can only predict one label per
segment, micro precision and recall are same as the percentage of correctly
classified segments.  Macro precision and recall are respectively the averages
of precision and recall for all classes. 
The optimal C value is determined separately for each of the algorithms by
cross-validation. 
 

Figure~\ref{fig:examplePCD} shows the original point cloud, ground-truth and predicted labels for one 
office (top) and one home scene (bottom). We see that on majority of the classes our model
predicts the correct label. It makes mistakes on some tricky cases, 
such as a pillow getting confused with the bed,
and table-top getting confused with the shelf-rack. 

One of our goals is to study the effect of various factors, and therefore
we compared various versions of the algorithms with various settings.  
We discuss them in the following.

\begin{table*}[tb!]
\vspace*{\captionReduceTop}
\caption{{\small Average micro precision/recall,  average macro precision and recall for home and office scenes. 
}}
\vspace*{\captionReduceBot} 
 \vskip -.05in
 \label{tbl:overall_result}
{\footnotesize
\newcolumntype{P}[2]{>{\footnotesize#1\hspace{0pt}\arraybackslash}p{#2}}
\setlength{\tabcolsep}{2pt}
\centering
\resizebox{\hsize}{!}
 {
\begin{tabular}
{p{0.23\linewidth}p{0.18\linewidth}|P{\centering}{12mm}P{\centering}{12mm}P{\centering}{12mm}|P{\centering}{12mm}P{\centering}{12mm}P{\centering}{12mm} }\\
\whline{1.1pt} 
& & \multicolumn{3}{c|}{Office Scenes} & \multicolumn{3}{c}{Home Scenes}  \\
\cline{3-8}
& & \multicolumn{1}{c}{micro} & \multicolumn{2}{c|}{macro} & \multicolumn{1}{c}{micro} &  \multicolumn{2}{c}{macro}   \\
\whline{0.4pt} 
     features &  algorithm & $P/R$ & Precision  & \multicolumn{1}{c|}{Recall} &  $P/R$ & Precision &  \multicolumn{1}{c}{Recall}  \\ 
\whline{0.8pt} 
None &  chance &  26.23 & 5.88 & 5.88 & 29.38 & 5.88 & 5.88\\
\whline{0.6pt} 
Image Only &           svm\_node\_only                        & 46.67  & 35.73 & 31.67 &   38.00 & 15.03 & 14.50\\
Shape Only &              svm\_node\_only                        & 75.36  & 64.56 & 60.88 &    56.25 & 35.90 & 36.52 \\
Image+Shape &         svm\_node\_only                         & 77.97  & 69.44 & 66.23 &   56.50 & 37.18 & 34.73 \\
\whline{0.6pt} 
Image+Shape \& context &  single\_frames                 & 84.32 & 77.84 & 68.12 & 69.13 & 47.84 & 43.62 \\
\whline{0.6pt} 
Image+Shape \& context &  svm\_mrf\_assoc   					& 75.94  & 63.89 & 61.79 &    62.50 & 44.65 & 38.34\\
Image+Shape \& context &  svm\_mrf\_nonassoc   					& 81.45  &76.79  &70.07   & 72.38  & 57.82  & 53.62 \\
Image+Shape \& context &  svm\_mrf\_parsimon	     			& 84.06  & 80.52  & 72.64   & 73.38  & 56.81  &54.80 \\

\whline{1.1pt} 
\end{tabular}
}
}
\vskip -.2in
\end{table*}

 \noindent
\textbf{Do Image and Point-Cloud Features Capture Complimentary Information?}
The RGB-D data contains both image and depth information, and enables us to
compute a wide variety of features.
In this experiment, we compare 
the two kinds of features: Image (RGB) and Shape (Point Cloud)
features.
To show the effect of the features independent of the effect of
context, we only use the node potentials from our model, referred to 
as \emph{svm\_node\_only}  in Table~\ref{tbl:overall_result}. The
\emph{svm\_node\_only} model is equivalent to the multi-class SVM formulation
\cite{joachims2009cutting}.  Table \ref{tbl:overall_result} shows that Shape
features are more effective compared to the Image, and 
the combination works better on both precision and recall.
This indicates that the two types of features offer complementary information and
their combination is better for our classification task.

\noindent
\textbf{How Important is Context?}
Using our svm\_mrf\_parsimon model as described in Section~\ref{sec:model},
we show significant improvements in the performance over using svm\_node\_only
model on both datasets. In office scenes, the micro precision increased by
6.09\% over the best svm\_node\_only model that does not uses any context. 
In home scenes the increase is much higher, 16.88\%.

The type of contextual relations we capture depend on the type of 
edge potentials we model.
To study this, we compared our method with models using only associative (svm\_mrf\_assoc)
or only non-associative (svm\_mrf\_nonassoc) edge potentials.
We observed that modeling all edge features using associative potentials 
is poor compared to our full model. In fact, using only associative potentials
showed a drop in performance compared to svm\_nodeonly model on the office
dataset.  This indicates it is important to capture the relations between
regions having \textit{different} labels.
Our svm\_mrf\_non\_assoc model does so, by modeling all edge features using
non-associative potentials, which can favour or disfavour labels
of different classes for nearby segments. It gives higher precision and recall
compared to svm\_nodeonly and svm\_mrf\_assoc. 




 However, not all the  edge features are  non-associative in nature, modeling
them using only non-associative potentials could be an overkill (each
non-associative feature adds $K^2$ more parameters to be learnt). Therefore
using our svm\_mrf\_parsimon model to model these relations achieves higher
performance in both datasets.

\noindent
\textbf{How Large should the Context Range be?}  
  \begin{wrapfigure}{r}{2in}  
 \vskip -.1in
 \includegraphics[width=2in]{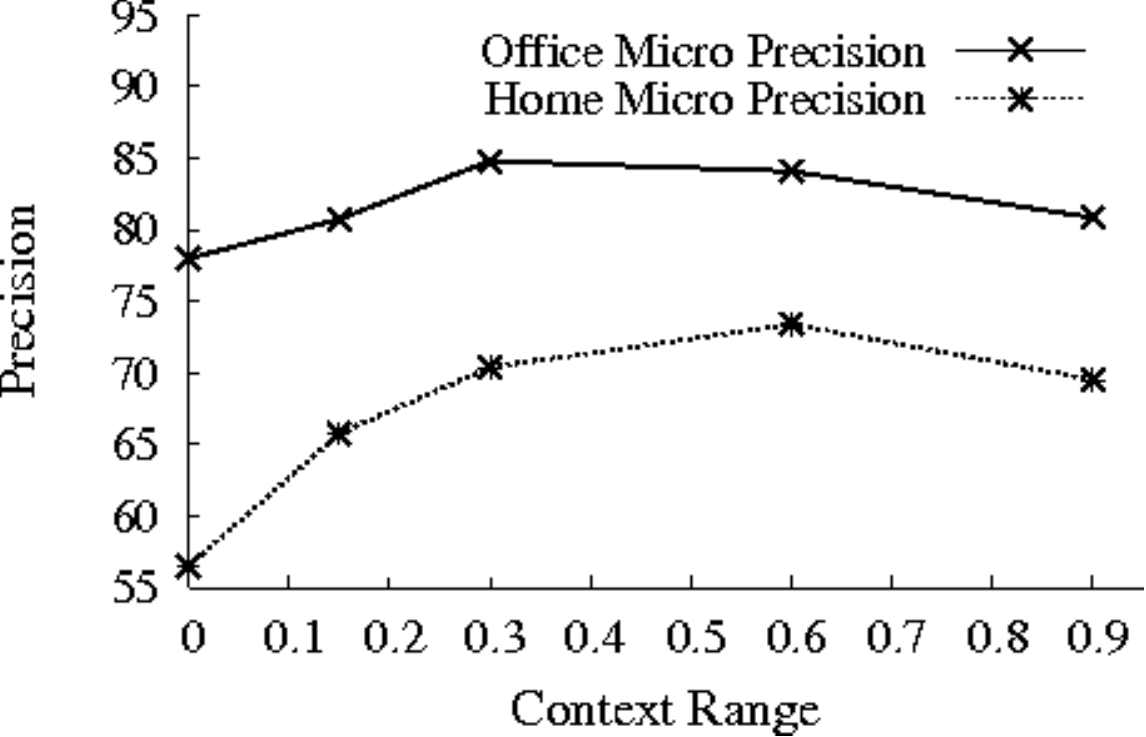} 
 \vspace*{\captionReduceTop}
 \vspace*{\captionReduceTop}
 \caption{{\small Effect of context range on precision (=recall here).}}
 \vspace*{\captionReduceBot}
 \label{fig:radiusplot}
 \end{wrapfigure}
Context relationships of different objects can be meaningful for
different spatial distances.  This range may vary depending
on the environment as well. For example, in an office, keyboard and
monitor go together, but they may have little
relation with a sofa that is slightly farther away.  In a house, 
sofa and table may go together even if they are farther away.

In order to study this, we compared our svm\_mrf\_parsimon with varying
context range for determining the neighborhood (see Figure~\ref{fig:radiusplot}
for average micro precision vs range plot). Note that the context range is determined
from the boundary of one segment to the boundary of the other, and hence it is
somewhat independent of the size of the object.  We note that increasing the
context range increases the performance to some level, and then it drops slightly.
We attribute this to the fact that with increasing the context range, irrelevant objects
may get an edge in the graph, and with limited training data, spurious
relationships may be learned.  We observe that the optimal context range
for office scenes is around 0.3 meters and 0.6 meters for home scenes.




\noindent
\textbf{How does a Full 3D Model Compare to a 2.5D Model?}
In Table~\ref{tbl:overall_result}, we compare the performance of our full model
with a model that was trained and tested on single views of the same
scene. During the comparison, the training folds were consistent with other
experiments, however the segmentation of this point-cloud was different (because
the input point-cloud itself is from single view). This makes the micro precision values not 
meaningful because the distribution of labels is not same for the two cases. In particular, 
many large object in scenes (e.g., wall, ground) get split up into 
multiple segments in single views. 
We observed that the macro precision and recall are higher when multiple views 
are combined to form the scene. 
We attribute the improvement in macro precision and recall to the fact that 
larger scenes have more context, and models are more complete because of multiple
views.



\noindent
\textbf{What is the Effect of the Inference Method?}
The results for svm\_mrf algorithms Table \ref{tbl:overall_result} were generated using the MIP solver. The graph-cut algorithm however, gives a higher precision and lower recall on both datasets. For example, on office data, the graphcut inference for our svm\_mrf\_parsimon gave a micro precision of 90.25 and micro recall of 61.74.  Here, the micro precision and recall are not same as some of the segments might not get any label.
Since it is orders of magnitude faster, it is ideal for realtime robotic applications.

%
%

   \vspace*{\subsectionReduceTop}
   \subsection{Robotic experiments}
   \vspace*{\subsectionReduceBot}
The ability to label segments is very useful for robotics applications,
for example, in detecting objects (so that a robot can find/retreive
an object on request) or for other robotic tasks such as manipulation.  We therefore
performed two relevant robotic experiments.

\noindent
   \textbf{Attribute Learning}: In some robotic tasks, such as
robotic grasping \cite{jiang_grasping} or placing \cite{jiang_placing}, 
it is not important to know the exact object
category, but just knowing a few attributes of an object may
be useful. For example, if a robot has to clean a floor, it would
help if it knows which objects it can move and which it cannot.
If it has to place an object, it should place them on horizontal
surfaces, preferably where humans do not sit. With this motivation we have
designed 8 attributes, each for the home and office scenes, giving a total of 10
unique attributes, comprised of: \emph{wall, floor, flat-horizontal-surfaces,
furniture, fabric, heavy, seating-areas, small-objects, table-top-objects,
electronics}. Note that each segment in the point cloud can have multiple
attributes and therefore we can learn these attributes using our model which
naturally allows multiple labels per segment.  We compute the precision 
and recall over the attributes by counting how many attributes were correctly
inferred. In home scenes we obtained a precision of 83.12\% and
70.03\% recall, and in the office scenes we obtain 87.92\%
precision and 71.93\% recall.

   \begin{figure}[t]
   \centering
 \includegraphics[width=0.9\linewidth,height=1.3in]{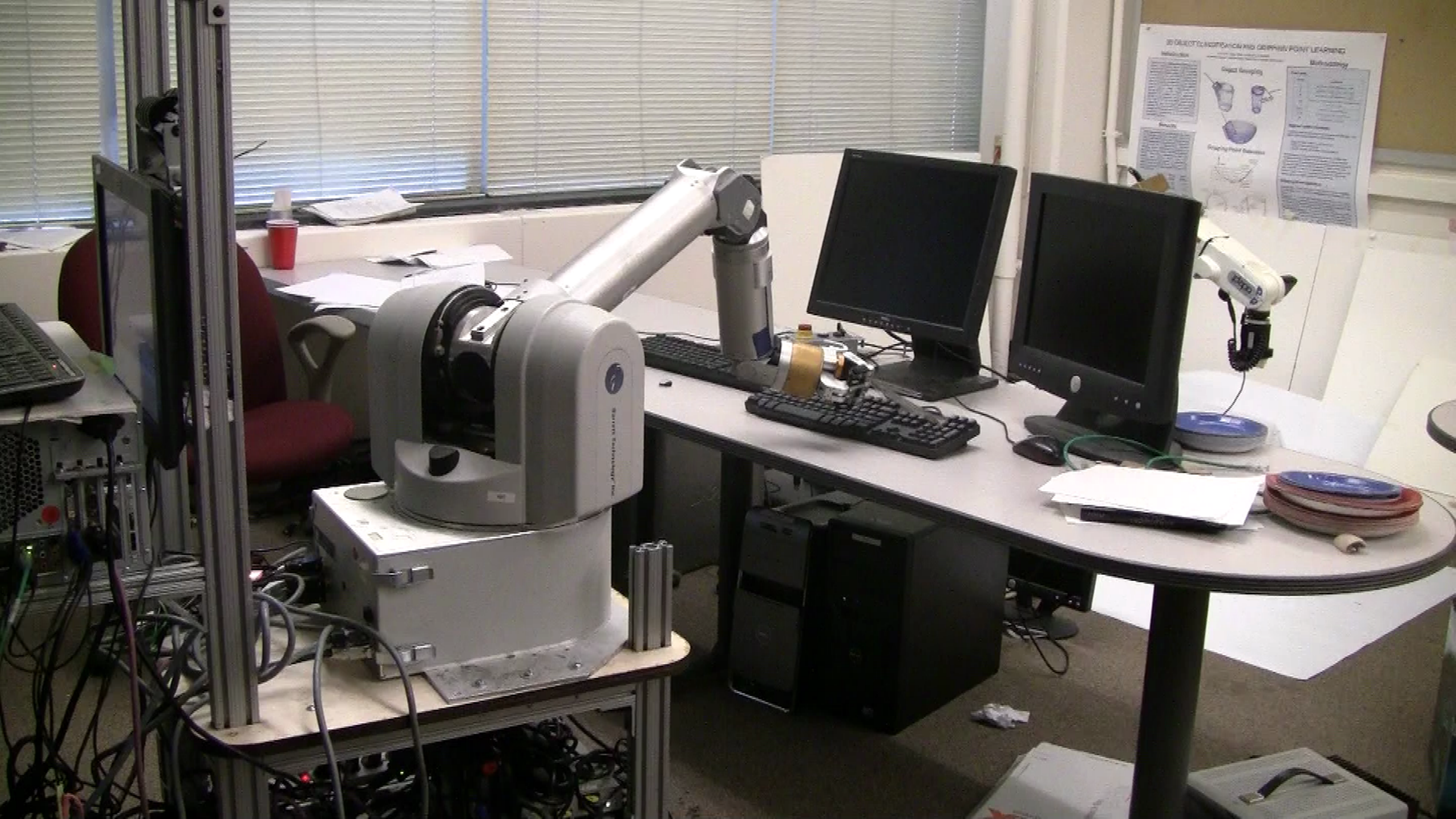} 
 \vspace*{\captionReduceTop}
 \caption{\small{Cornell's POLAR (PersOnaL Assistant Robot) using our classifier for detecting a keyboard in a cluttered room.}}
 \vspace*{\captionReduceBot}
 \vskip -.04in
 \label{fig:robot}
 \end{figure}

\noindent
\textbf{Robotic Object Detection}: 
We finally use our algorithm on a mobile
robot, mounted with a Kinect, for completing the goal of finding 
an object such as a keyboard in an extremely cluttered room (Fig.~\ref{fig:robot}).  
The following video 
shows our robot successfully
finding the keyboard in an office:
   \texttt{http://pr.cs.cornell.edu/sceneunderstanding}

\smallskip
In conclusion, we have proposed and evaluated the first model and learning 
algorithm for scene understanding that exploits rich relational information
from full-scene 3D point clouds. We applied this technique to object 
labeling problem, and studied affects of various factors on a large dataset.
Our robotic applications shows that such inexpensive RGB-D sensors can
be quite useful for scene understanding by robots.

\smallskip
\noindent
\textbf{Acknowledgements.} We thank Gaurab Basu, Matthew Cong
and Yun Jiang for the robotic experiments.


{\small
\bibliographystyle{IEEEtran}
\bibliography{references}
}

\end{document}